# Recent advancement in Disease Diagnostic using machine learning: Systematic survey of decades, comparisons, and challenges

**Farzaneh Tajdini [1, *], Mohammad-Javad kheiri [2]**

[1] Department of Computer Engineering, Malek Ashtar University, Tehran, Iran
[2] Department of Financial Management, Arak Branch, Islamic Azad University, Arak, Iran

**Highlights**

- ML techniques have been widely used in the literature as fast, affordable, and non-invasive approaches for CAD prediction.
- This paper conducts a comprehensive review of all relevant studies between 1992 and 2019 for ML-based Diagnostic using machine learning.
- The impacts of dataset characteristics and applied ML techniques are investigated in detail.

**ABSTRACT**

Computer-aided diagnosis (CAD), a vibrant medical imaging research field, is expanding quickly. Because errors in medical diagnostic systems might lead to seriously misleading medical treatments, major efforts have been made in recent years to improve computer-aided diagnostics applications. The use of machine learning in computer-aided diagnosis is crucial. A simple equation may result in a false indication of items like organs. Therefore, learning from examples is a vital component of pattern recognition. Pattern recognition and machine learning in the biomedical area promise to increase the precision of disease detection and diagnosis. They also support the decision-making process's objectivity. Machine learning provides a practical method for creating elegant and autonomous algorithms to analyze high-dimensional and multimodal bio-medical data. This review article examines machine-learning algorithms for detecting diseases, including hepatitis, diabetes, liver disease, dengue fever, and heart disease. It draws attention to the collection of machine learning techniques and algorithms employed in studying conditions and the ensuing decision-making process.

## 1. INTRODUCTION

The rapid emergence of enormous amounts of healthcare data will radically alter how medical treatment is provided. Many patients' treatments will be delivered primarily through doctor-patient interaction, strengthened by new insights from machine learning. [68] Machine learning is advantageous because it eliminates the need for explicit human characterization by allowing significant links between different pieces of data to be learned from the data. The ability of a machine learning-based technique to consider a larger variety of facts than a doctor may, at its most basic level, result in a more accurate diagnosis. However, when taken a step further, it allows for a better understanding of the disease ("Big Data") by highlighting the patterns in the data. From a scientific/medical research perspective, machine learning is also likely to play a role in aiding clinicians in delivering care to patients.[69]

Machine learning is particularly well suited to classification tasks, such as medical image recognition, where the input is a digital image, and the output is binary ("normal" or "disease"). When identifying previously unseen photos of biopsy-validated lesions, for instance, AI has proven to be more sensitive and specific than dermatologists after initial training in classifying suspected skin lesions as either benign or malignant with input from dermatologists. [70]

One advantage of applying AI to the analysis of medical tests, such as imaging, is that it makes it easier to evaluate tests performed in remote or underserved areas. This can lead to an accurate and timely diagnosis and, if necessary, the ability to refer a patient to expert care at an earlier stage, potentially changing the course of the disease. For instance, remote clinics in many countries with a high tuberculosis prevalence lack radiological expertise [71]. A recent study involving such a system reported that AI (which had been pretrained using active pulmonary tuberculosis images confirmed by a cardiothoracic radiologist) correctly diagnosed active pulmonary tuberculosis with a sensitivity of 97.3% (sensitivity = true positive/[true positive + false negative]) and specificity of 100% (specificity = true negativity). However, radiographs remotely uploaded from these centers can be interpreted by a single central system using AI. [72] Additionally, if the volume of medical imaging grows faster than the supply of skilled radiologists, AI will play an increasingly significant role in picture interpretation.

## 2. Machine learning

ML is the process of teaching a computer to use its prior knowledge to address a problem that has been presented to it. Due to the current accessibility of less expensive processing power and memory, the idea of applying ML in several fields to solve problems faster than humans has attracted substantial interest. This enables the processing and analysis of extraordinarily vast amounts of data to find insights and correlations between the data that are not immediately apparent to the human eye. Its intelligent behavior is built on many algorithms that allow the computer to create salient judgments by abstracting knowledge from experience. On the other hand, the more sophisticated approach of deep learning (DL) will enable computers to automatically extract, evaluate, and comprehend the relevant information from the raw data by modeling how people think and learn [10]. Specifically, a group of methods known as deep learning is neural data-driven and rely on autonomous feature engineering procedures. Its accuracy and high performance are due to the automatic learning of input features [10]. In ML and DL, the classification algorithm determines whether or not a judgment will be successful. Different classification methods are available in ML, and their performance is respectable. These algorithms were created specifically for classification purposes. Even while ML performs pretty well, DL is currently taking place in most classification applications. The main distinction between ML and DL is the method used to extract the classifier's characteristics. The classification performance of DL, which draws its features from several non-linear hidden layers, is significantly better than that of ML, which depends on manually created features.

### 2.1. Classifiers

To make raw data machines understandable and ready for feature extraction, data evaluated under ML/DL must undergo several preprocessing procedures. Collected data is analyzed based on a set of qualities known as features. The features under consideration must be non-redundant and able to differentiate. In this approach, training time and difficulties with overfitting are reduced. There are numerous ways to extract characteristics. The data can be labeled after the features have been removed. The process through which a machine labels data is referred to as a classifier. In other words, a device classifies data using various classifier techniques. The most popular classifiers include SVM, RF, LR, DT, NB, KNN, and others.

On the other hand, DL creates a comprehensive network modeled after a biological neural network to carry out the complete ML process, as opposed to the step-by-step method used by ML. It employs many nonlinear processing unit layers. The following unit receives the output of the previous unit as input. Each level of the hierarchical structure of data movement converts the data it gets into more abstract data so that the next level can be fed.

**Table 1.** Key Concepts in Machine Learning

| Concepts | Definition |
| --- | --- |
| Supervised, unsupervised, and semi-supervised learning | Based on prior data that contains labels or classes, supervised learning predicts labels or courses on future data. Unsupervised learning finds structure in unlabeled data, typically in the form of clusters. Humans assign labels to the systems discovered by unsupervised learning once semi-supervised learning has been completed. |
| Classification and regression | Both are supervised learning methods. Classification predicts discreet categories, such as normal versus diseased, while regression predicts real-valued outputs, such as response to therapy. |
| Ensemble learning | The average of all the models is used by ensemble methods to create predictions. Random forests, gradient-boosting, and stacking/meta-ensembles are typical ensemble methods. |
| Deep learning | artificial neural networks with multiple layers that can learn sophisticated nonlinear functions. Very helpful for unstructured data like pictures, audio, or text but usually does not give any insights into the data characteristics that power the operations. |
| Bayesian learning | Methods that combine prior knowledge in addition to data to perform machine learning. |
| Dimensionality reduction | Reduces a dataset's number of attributes or features by selecting important features or combining elements to capture variance in a dataset. They often improve machine learning models' performance and aid visualization. |

| Federated learning | techniques for learning progressively from data dispersed across numerous locations and cannot be merged into a single dataset. Federated learning is advantageous when learning from sensitive personal data or when data are spread across several clinical systems. |
|---|---|

## 3. Use of AI In Medicine

### 3.1. Diagnosis of Diseases by Using Different Machine Learning Algorithms

Many studies have focused on different machine-learning methods for disease diagnosis. Machine-learning algorithms have been proven to be effective in the diagnosis of a variety of ailments by researchers. The ailments identified by MLT in this survey article include hepatitis, dengue, diabetes, liver, and heart disease.

#### 3.1.1. Alzheimer's disease

To make an early AD prediction, a computer method utilizing an SVM-based ML approach was looked into in [11], where the gene-protein sequence was considered a potential data source. It was claimed that an ML-based strategy could be a promising way to predict AD by utilizing the sequencing data of gene-coding proteins based on the acquired classification performance. In [12], an ML model for early AD diagnosis was put up, using voice processing to extract numerous linguistic aspects. Different feature selection strategies were used to further process the extracted linguistic features (syntactic, semantic, and pragmatic) from 242 affected and 242 unaffected persons with AD. The ML classifier is then fed the chosen features. The proposed ML model successfully separates AD patients from NC with a precision of 79%, utilizing KNN feature selection with an SVM classifier. The authors of [13] looked at a machine-learning prediction model for early AD detection based on patient-specific neuropathological abnormalities. Clinical symptoms were seen as less explicit and certain in this case than post-mortem neuropathological abnormalities. The authors hypothesized that the presented model would not be suitable for clinical application, but it could be a step toward precision medicine in AD, given the obtained accuracy of 77%. In [14], ML was used to distinguish between age-matched cohorts of 48 AD, 75 EMCI, 39 LMCI patients, and 51 NC. With KNN feature selection and an SVM classifier, the suggested ML model distinguishes AD patients from NC with a maximum accuracy of 79%. According to patients' neuropathological abnormalities, the authors of [13] looked into a machine-learning prediction model for early AD detection. In this case, clinical symptoms were not seen as explicit and were certain as post-mortem neuropathological alterations. However, the scientists indicated that, despite the obtained accuracy of 77%, the proposed model would not be suitable for clinical use but could nonetheless be a step toward precision medicine in AD. In [14], the age-matched 48 AD, 75 EMCI, 39 LMCI, and 51 NC patients were distinguished using machine learning (ML). From the four separate groups—AD, stable MCI (sMCI), progressive MCI (pMCI), and NC—with 92, 82, 95, and 92 patients each—a total of 361 subjects were chosen for the experiment. It was demonstrated that the suggested strategy outperformed various common ML algorithms regarding classification accuracy.

### 3.2. Machine Learning and Deep Learning for Heart Failure Risk Prediction

The accurate implementation of numerous therapy alternatives, from pharmacological to invasive mechanical ventricular support and heart transplantation, depends on the precise prognosis of HF outcome. Recent HF outcome prediction studies have mostly used wearable sensors, EHR, and echocardiographic data. In one study, domain experts and data-driven selected features were combined in machine learning models, including decision tree models, using quantitative characteristics derived from echocardiogram images. The prediction accuracy of data-driven feature selection was significantly higher than that of expert-driven feature selection [16]. In a different investigation, the timing and amplitude of left ventricular (LV) pictures were examined to learn more about the information on myocardial mobility and deformation in the cardiac cycle under rest and stress. Their results suggested that LV images provide informative features for HF with preserved ejection fraction (HFpEF) prediction [17].

### 3.3. Machine Learning and Deep Learning for Prediction of Hospital Readmissions

The care of HF patients has several difficulties due to the high hospital readmission rate. It is commonly acknowledged that HF patients with an increased risk of readmission can be precisely identified and targeted with management algorithms, improving HF outcomes and healthcare costs. Most machine learning techniques use EHR data to identify HF patients at high risk for readmission. In one investigation, the naive Bayes model was applied to patient primary encounter data to predict HF readmission [18]. For each subset of the patient cohort, the top related features in HF readmission were chosen and integrated as the input of the predictive naive Bayes model. A tree-based model, adopted from the random forest model, was proposed to predict HF readmission using demographic, socioeconomic, utilization, service-based, comorbidity, and severity features.

Deep learning models were also suggested for HF prediction and conventional machine learning methods. A multi-layer perception (MLP) model was created in one study to forecast HF readmission using EHR data [22], which included demographics, admission features, medical history, visits to emergency rooms, history of medication usage, and healthcare services received outside of the hospital. The readmission rate is uneven, but the MLP model accepts it because it is modest compared to most patients (who are not readmitted). Another study employed the deep, unified network (DUN) model to predict HF readmissions. This model integrates the output from each hidden layer to capture potentially useful information [23]. The DUN model surpassed the gradient boosting, logistic regression, and conventional deep neural networks. The long short-term memory model has also been effectively used for HF outcome prediction to deal with the longitudinal temporal data of EHR [22]. Telemonitoring data have also been utilized to identify HF patients at high risk of readmission in addition to EHR data [19]. The ability to predict HF readmission using EHR, tele-HF and wearable sensor data is very promising.

### 3.4. Machine Learning and Deep Learning for Mortality Prediction

For HF patients to make optimal therapy decisions, accurate mortality prediction is essential. Due to the lack of consistent marker factors, data noise, and unbalanced data sets, this prediction is difficult to make. One study employed EHR data to predict HF mortality using machine learning techniques, including random forests, logistic regression, AdaBoost, decision trees, and support vector machines [24]. A comprehensive set of data, including all baseline demographic, clinical, laboratory, electrocardiographic, and symptom data, was used to evaluate a similar set of machine learning models for mortality prediction of HF patients [25]. Among these models, the random forest model performed the best.

Other recent studies have applied deep learning algorithms to enhance mortality prediction in HF patients. In one study, the most crucial features were mined to create a unique deep learning model called Feature Rearrangement-based Convolution Net (FReaConvNet), which was then used to predict in-hospital, 30-day, and 12-month mortality. Enhancing prediction accuracy in unbalanced data sets requires feature importance analysis. Machine learning analysis of EHR and laboratory data identified serum creatinine and ejection fraction as crucial characteristics [26]. Compared to using all other evidence of EHR data, employing just these two factors improved prediction results [26]. In another study, novel and complex computer vision models using convolutional networks to calculate heart motion trajectories accurately predicted patient survival using 4D imaging of the heart (3D MRI images + time) [27].

### 3.5. Machine Learning and Deep Learning for Skin cancer

One of the diseases that is seemingly spreading quickly over the world is skin cancer. Skin cancer is a condition in which abnormal skin cells develop out of control [[28], [[29], [[30], [[31], [32,33,34]. Melanoma and non-melanoma are the two forms of skin cancer that develop. Non-melanoma, a deadly form of cancer or tumor that starts on the skin and resembles a dark spot, is less dangerous than melanoma. In some cases, they may develop from a mole to a growth in size, irregular edges, itching, skin breakdown, and colour difference. Three thousand three hundred thirty females and 5990 males died from skin cancer in 2018, accounting for 9320 fatalities [35]. Asia, Latin America, and Africa have lower rates of melanomas than North America, Europe, Australia, and New Zealand. The study found that early skin cancer detection can reduce death rates by up to 90%. Therefore, early cancer detection is important and could aid in lowering the associated risk factors [36].

### 3.6. Machine Learning and Deep Learning for Lung cancer

One of the most common causes of lung death worldwide is lung cancer. Several patients will benefit from operational, percutaneous, and surgical treatments if the issue is a small, diffuse tumor. In 75 percent of lung cancer cases, a diagnosis of progressive clinical illness, nodal progression, and metastatic disease emerge later due to the rarity of those without any symptoms in the early stages of the disease. According to Australian studies, the total survival rate of people with lung cancer is 15% [37]. Several researchers have used the LIDC/IDRI Database to report on their work in lung nodule identification and categorization. The database contains more than 240000 photos of nodules.

### 3.7. Machine Learning and Deep Learning for Breast cancer

Breast cancer, which develops in breast cells, is the second most common cancer in women worldwide after skin cancer. Breast cancer can affect both men and women, but it is significantly more common in women [38]. The most recent advancement in medical imaging for the early detection of breast cancer is machine-assisted systems, which also improve radiologists' diagnostic abilities. The most often used techniques for detecting breast cancer include mammography, tomography, breast ultrasound (BUS), MRI, and CT scans, while PET is also recommended [39]. Only a few of these operations are typically indicated because the breast is considered the most oversensitive organ in the human body, depending on the patient's state and the tumor's status. At an early stage of breast cancer, mammography is regarded as a low-cost and secure technique, although it is useless in young females with dense breasts. The BUS procedure is thought to support mammograms [40] to avoid unnecessary biopsies. Publically accessible breast imaging datasets include DDSM, MIAS, WBCD, BCDR, and NBIA [41]. Before segmentation, pre-processing techniques such as pectoral muscle removal and artefact reduction are carried out after image acquisition. The machine-assisted system's most crucial stage for improving accuracy and lowering false positives for the presence of abnormality is segmentation [42]. The GLCM approach was advocated by many studies [43,44,45] to define texture-based features. Like the previous method, LBP is a wonderful tool for separating benign from malignant tumors using texture extraction [46].

**Table 2.** Example Applications of Machine Learning for Diagnosis and Treatment.

| Dataset | Goals | Successes | Data Type | ML Method |
|---|---|---|---|---|
| Patient molecular patient profiles without clinical data | (1) Discover subtypes or stratify patients; (2) Identify similarities among clustered patients | Cancer subtyping (Curtis et al., 2012, [1] Gao et al., 2019) [2] | High-dimensional, structured data; Unlabeled data | Unsupervised clustering for cluster discovery; supervised learning and deep learning for subtype assignment. |
| Patient or laboratory molecular profiles with clinical data | Predict the most efficacious therapies | Cancer cell line drug response prediction (Chiu et al., 2019, [3] Costello et al., 2014) [4] | High-dimensional, structured data; Unlabeled data | Supervised learning, deep learning, and ensemble learning |
| Images and associated diagnoses | Automated diagnoses | Medical imaging diagnostics (Liu et al., 2019) [5] | Unstructured data; Labeled data | Deep learning |
| EMR data + clinical outcomes | Predict clinical outcomes | Diagnosis of gestational diabetes (Artzi et al., 2020) [6]; patient similarity (Lee et al., 2018) [7] | Structured and unstructured data; Labeled data | Traditional machine learning on structured data with labels; deep learning/natural language processing to mine unstructured data; federated learning |
| Wearable and home device ambient data collection | Early diagnoses | Detection of atrial fibrillation (Bumgarner et al., 2018) [8] and agonal breathing, an audible biomarker of cardiac arrest (Chan et al., 2019) [9] | Unstructured, longitudinal data; Labeled data | Both supervised and deep learning approaches, with adjustments made for time-series analyses. |
| Deep longitudinal data | Ongoing health management | None yet due to a lack of available datasets | Structured and unstructured data; Labeled data | Continuous learning |
| Saba, [47] | Multiple classifiers voting | LIDC | 100 (Sensitivity) | |
| Shen et al., [48] | Multi-scale Convolutional Neural | LIDC-IDRI | 86.84 (Accuracy) | |

| Author | Method | Dataset | Results |
|---|---|---|---|
| | Networks (MCNN) | | |
| Kumar et al., [49] | deep feature with auto-encoder | LIDC | 75.01 (Accuracy) 83.35 (Sensitivity) 0.39 (false positive) |
| Firmino et al., [50] | Watershed, HoG, and SVM | LIDC-IDRI | 97 (Accuracy) 94.4 (Sensitivity) 7.04 (false positive) |
| Setio et al., [51] | multi-view ConvNet | LIDC-IDRI | sensitivity of 85.4% and 90.1% at 1 and 4 FPs/scan |
| Saba et al., [52] | Deep convolutional neural network (DCNN) | PH2, ISBI 2016, and ISBI 2017 | 98.4% on the PH2 dataset, 95.1% on the ISBI dataset, and 94.8% on ISBI 2017 dataset |
| Ramya et al., [53] | Used active contour segmentation mechanism. GLCM feature, and for the classification, they used the SVM classifier | ISIC | 95% (Accuracy) 90% (sensitivity) 85% (specificity) |
| Premaladha and Ravichandran [54] | Median filter and Contrast Limited Adaptive Histogram Equalization and Normalized Otsu's Segmentation are used. neural networks and hybrid AdaBoost SVM are used for the classification | Skin Cancer and Benign Tumor Image Atlas - Contents | 91.7% (Accuracy) 94.1%(sensitivity) 88.7%(specificity) 0.83%(Kappa) |
| Bareiro Paniagua et al., [55] | The feature is extracted using the ABCD rule, and the extracted feature is classified using the SVM | PH2 | 90.63% (Accuracy) 95% (sensitivity) 83.33%(specificity) |
| Patel and Mishra, [56] | Used k-state grouping, histogram alignment, and the zack, SVM is used for the classification | | 93.57 (Accuracy) |
| Kazemi et al., [57] | Used SVMs to classify acute myelogenous leukemia. | | 96% (Accuracy) 95% (sensitivity) 98% (specificity) |
| Khalilabad and Hassanpour [58] | J48 Tree | | 95.45 (Accuracy) |
| Rehman et al., [59] | CNN using AlexNet architecture | | 97.78 (Accuracy) |
| Sharma and Kumar, [60] | PCA based ABCBPNN | | 98.7%(Accuracy) 0.478% (FAR) 0.398% (FRR) |
| Zhang et al., [61] | CNN features as the input of SVM for classification Combined CNN and HOG features | | 95.9% (Accuracy) 96.1% (sensitivity) 94.57% (specificity |
| Abdel-Zaher and Eldeib, [62] | Deep belief network unsupervised path followed by backpropagation supervised path | WBCD | 99.68 (Accuracy) 100 (sensitivity) 99.47 (specificity) |
| Sun et al., [63] | Semi-supervised learning with Convolution neural network | FFDM | 82.43 (Accuracy) 81.00 (sensitivity) 72.26 |

| Sadad et al., [64] | Fuzzy C-Means and region-growing based technique for segmentation, LBP-GLCM, and LPQ technique used for feature extraction | MIAS ———— DDSM | 97.2 (Accuracy) 97 (Specificity) 98 (Sensitivity) 97 (F-Score) 94 (MCC) |
|---|---|---|---|
| Li, [65] | CNN, Adaboost, RF, SVM | CT | 80.06 (DSC) 82.67 (precision) 84.34 (recall) |
| Ben-Cohen et al., [66] | Fully convolutional network (FCN) | CT | 0.86 (true positive rate) 0.6 (false positive per case) |
| Chang et al., [67] | texture, shape, kinetic curve, and logistic regression analysis were used for the classification | CT | 81.69 (Accuracy) 81.82 (sensitivity) 81.63 (specificity) |

## 4. Discussion

Analysis of medical images and healthcare data, including whole-slide pathology [79], X-ray [80], diabetes [81, 82], breast cancer [83], heart [84], time series [85], medicinal plants [86], stock market [87], stroke [88], etc., has successfully used machine learning and deep learning [77]. According to the authors, AI accomplished this goal by mapping the complete input data from many known disease cases into a high-dimensional space. Then, when the data from a new patient was given, AI mapped it into the same high-dimensional space and classified the patient. The authors concluded that while pulmonologists' interpretations of pulmonary function tests were prone to significant variations and errors, those provided by AI-based software were more accurate (and consistent). They could be used as a potent decision-support tool to enhance clinical practice. [74] However, it has been mentioned elsewhere that real clinicians may not have performed as well as they should have because they had little access to clinical data [74]. Whether or whether the clinicians' performance was overestimated, this study demonstrated that AI has potential applications in respiratory medicine beyond picture analysis. Furthermore, physicians and researchers in the field of primary care believe there is a critical need for quick answers in one of the areas of respiratory research: diagnosis. [75] The International Primary Care Respiratory Group repeatedly discovered the need for "simple tools" (such as questionnaires) to facilitate illness detection and assessment in community settings during a prioritization exercise [75].

These results imply that AI/machine learning offers a novel method for creating diagnostic algorithms that may help differentiate and diagnose respiratory ailments, among other disorders and diseases. People tend to be more linear in their associations, which is why people cannot analyze numerous issues at once as AI-based algorithms can, which accounts for AI's greater speed/efficiency. The US FDA has authorized several AI-based medical algorithms (Table I) [76].

## 5. Conclusion

The assessment field has been inundated by statistical models for estimating that cannot yield good performance results. Large data points, missing values, and categorical data cannot be held in statistical models. These and other factors highlight the value of MLT. Many applications, including image detection, data mining, natural language processing, and illness diagnostics, rely heavily on machine learning. ML suggests potential answers for each of these fields. This paper offers an overview of machine learning algorithms for diagnosing various disease. Numerous algorithms have produced positive outcomes because they correctly detect the feature. The survey outlines the benefits and drawbacks of various algorithms. Detailed improvement graphs of machine learning algorithms for disease prediction are shown. Analysis shows that these algorithms offer improved accuracy for diagnosing various diseases. This survey article also includes a group of tools created by the AI community. These tools are very useful for the analysis of such problems and provide an opportunity for an improved decision-making process.

**Contributions**

The following is a summary of this survey's significant contributions:
- To find the most precise method of detection, we have combined recent studies on three brain illnesses (such as Skin cancer, Lung cancer, and Breast cancer) that leverage ML and DL.
- A succinct introduction of the most popular feature extraction techniques for diagnosing cancer is given.
- The main conclusions from the reviewed articles are then distilled. Additionally, several open problems and potential study topics are offered.